# SEQUENTIAL CROSS ATTENTION BASED MULTI-TASK LEARNING

*Sunkyung Kim* *Hyesong Choi* *Dongbo Min**

Department of Computer Science and Engineering, Ewha Womans University, Seoul, Korea
E-mail: skk095@ewhain.net, hyesongchoi2010@ewhain.net, dbmin@ewha.ac.kr

## ABSTRACT

In multi-task learning (MTL) for visual scene understanding, it is crucial to transfer useful information between multiple tasks with minimal interferences. In this paper, we propose a novel architecture that effectively transfers informative features by applying the attention mechanism to the multi-scale features of the tasks. Since applying the attention module directly to all possible features in terms of scale and task requires a high complexity, we propose to apply the attention module sequentially for the task and scale. The cross-task attention module (CTAM) is first applied to facilitate the exchange of relevant information between the multiple task features of the same scale. The cross-scale attention module (CSAM) then aggregates useful information from feature maps at different resolutions in the same task. Also, we attempt to capture long range dependencies through the self-attention module in the feature extraction network. Extensive experiments demonstrate that our method achieves state-of-the-art performance on the NYUD-v2 and PASCAL-Context dataset. Our code is available at https://github.com/kimsunkyung/SCA-MTL

## 1. INTRODUCTION

Convolutional neural networks (CNN) have significantly improved the performance of scene understanding tasks from visual data, including monocular depth estimation [1, 2, 3] and semantic segmentation [4, 5]. While most approaches focus on advancing the performance of a single task, some approaches have attempted to infer multiple tasks within a single network [6, 7, 8]. These multi-task learning approaches would be essential in deploying a visual scene understanding system, which often requires inferring both geometric and semantic cues from scenes.

The multi-task learning approaches [7, 6, 9, 10] mainly focus on investigating how to transfer useful information between multiple tasks. In [10], they proposed a cross stitch unit to automatically learn an optimal combination of shared representations over multiple tasks. The method in [9] proposed a new architecture, termed Multi-Task learning with Attention Network (MTAN), that forms task-shared and task-specific networks using a task attention module. However, these approaches [10, 9] may face a huge computational overhead issue as the sub-networks increase linearly in proportion to the number of tasks without sharing encoders for feature extraction. To alleviate this problem, some recent works [6, 7, 11] proposed to share a backbone network and designed task-specific heads and modules with the purpose of exchanging information across tasks. In [6], they proposed multi-task distillation module which refines the features of the current task by using other task features. In [7], they introduced an efficient method to improve an individual task prediction by capturing cross task contexts. A modern Neural Architecture Search (NAS) method [12] was employed to automatically find an optimal context type from five context type candidates (global, local, T-label, S-label, and none). While these methods [6, 7] consider interactions between the task features within the same scale only, the method in [11] attempts to leverage interactions between the task features on multiple scales. To this end, the multi-task distillation is applied on multiple scales by noting that each task can work in different receptive fields. In addition, they proposed a feature propagation module (FPM) that propagates information among multi-scale feature maps. However, this method [11] does not fully utilize information of tasks on different scales.

In this paper, we propose a novel architecture that effectively transfers informative features of different scales among tasks through the attention module. A target task feature can be augmented by extracting useful information on features with various resolutions of multiple tasks. However, since a straightforward application of the cross-attention module (CAM) to all possible features in terms of scale and task requires a prohibitively high complexity, we propose to apply the CAM sequentially for the scale and task. To be more specific, for a target task, the features extracted on the same scale from source tasks are used to augment the target feature using the cross-task attention module (CTAM). The task-augmented features are then progressively refined using the cross-scale attention module (CSAM) where the features within the same task are upsampled with the guidance of their fine features. Additionally, motivated by recent single task approaches [13, 14] that simultaneously leverage the convolutional layers and self-attention modules for capturing long range dependencies and modeling local features well, we attempt to boost the feature maps from convolutional backbone networks with the self-attention module. Our main contributions are summarized as follows.

1. We propose a new method for applying the CAM to the multi-task learning architecture in terms of the task (CTAM) and scale (CSAM) for augmenting the task features by extracting useful information on features with various resolutions of multiple tasks.

2. For the multi-task learning model, we are the first attempt to boost the convolutional features from the backbone network by capturing the long range dependencies through the self-attention module.

3. Extensive experiments conducted on the NYUD-v2 and Pas-

* Corresponding author
This work was supported by the Mid-Career Researcher Program through the NRF of Korea (NRF-2021R1A2C2011624). Sunkyung Kim is grateful for financial support from Hyundai Motor Chung Mong-Koo Foundation.

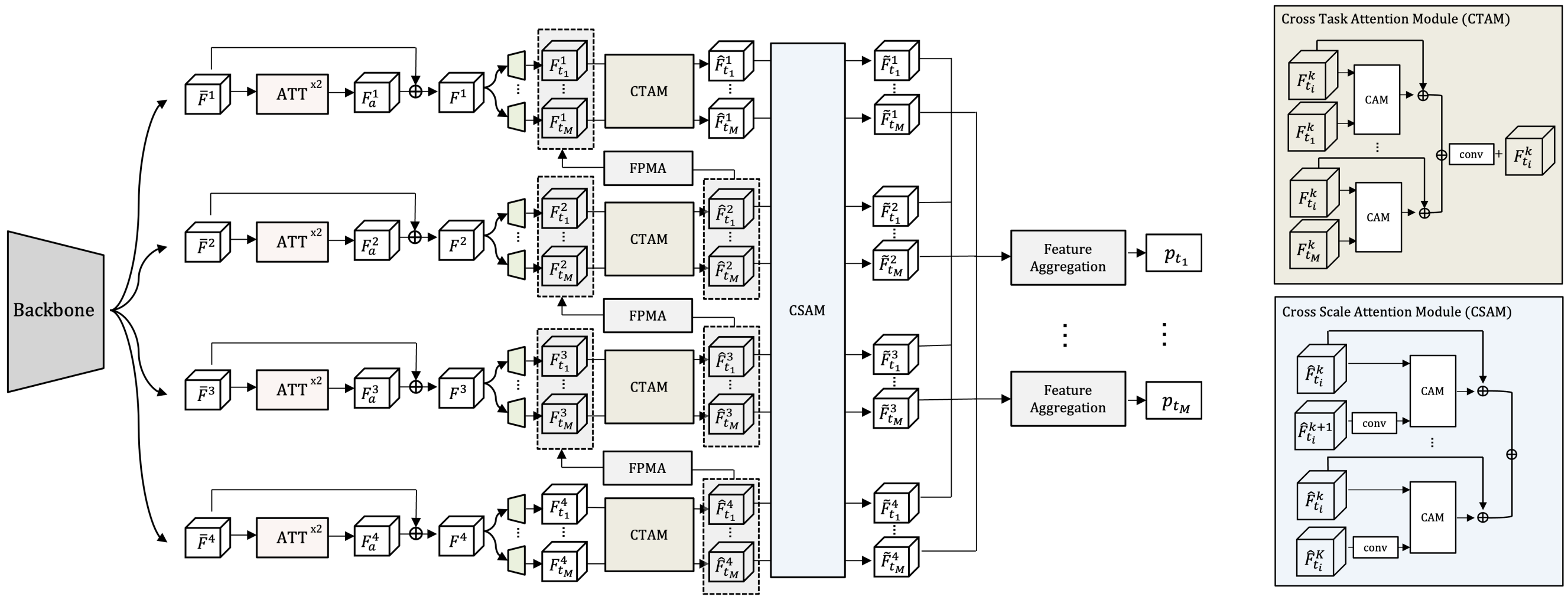

**Fig. 1**: Our network architecture. The attention-augmented feature map $F^k$ is generated by concatenating the convolutional feature map $\overline{F}^k$ and the attention feature map $F_a^k$ on the scale $k = 1, ..., K$ where $K$ is the number of scales. Here, $K$ is set to 4. The attention-augmented feature map is put into the task-specific head to obtain the task-specific feature map $F_{t_i}^k$ for $i = 1, ..., M$ where $M$ indicates the number of tasks. $\hat{F}_{t_i}^k$ is generated through the CTAM that applies the cross-attention between the different tasks at the same scale $k$. Also, a low-resolution scale information from the CTAM is transmitted to a high-resolution scale through the FPMA. After that, $\hat{F}_{t_i}^k$ is put into the CSAM for imposing the cross-attention across different scales, generating the final feature map $\tilde{F}_{t_i}^k$

cal Context datasets demonstrate significant performance improvement compared to the recent multi-task learning approaches.

## 2. PROPOSED METHODS

### 2.1. Overview and architecture design

Fig. 1 depicts the overall framework of the proposed multi-task learning model based on the attention modules. Multi-scale convolutional features extracted from the convolutional backbone networks (e.g., HRNet-18 [15]) are first passed through the self-attention module. To preserve the local details of the convolutional feature maps while capturing long range dependencies, we employ the attention module of swin transformer [16] designed for dense prediction tasks. The self-attention module is applied twice on each scale. Formally, the attention-augmented feature map $F^k$ is generated by concatenating convolutional feature map $\overline{F}^k$ and the attention feature $F_a^k$ where $k$ represents a scale. Here, we use four different feature maps (1/4, 1/8, 1/16, 1/32) from the backbone networks. The attention-augmented feature maps, which are shared with all tasks, are then put into task-specific heads for generating the task-specific features $F_{t_i}^k$ for $i = 1, \ldots, M$ where $t_i$ and $M$ indicate a task index and the number of tasks, respectively. The task-specific features on the scale $k$ are augmented using the CTAM such that for target task feature $F_{t_i}^k$, the key and value from a source task feature $F_{t_j}^k$ for $j = 1, \ldots, M$ (except $j \neq i$) are used for performing the cross-attention. Note that the feature propagation module with attention (FPMA) is used to transfer feature information of the CTAM output feature map $\hat{F}_{t_i}^k$ into the next scale. These feature maps are then put into the CSAM for imposing the cross attention across different scales, generating a final feature $\tilde{F}_{t_i}^k$. The prediction map $p_{t_i}$ of the $i^{th}$ task is finally obtained via the feature aggregation that fuses the feature maps $\tilde{F}_{t_i}^k$ for $k = 1, \ldots, 4$ over all scales.

### 2.2. Cross Task and Cross Scale Attention Modules

The CTAM and CSAM aim to perform the cross-attention across tasks and scales, respectively. While the CTAM is applied to the feature maps of multiple tasks on the same scale, the CSAM is for the feature maps of the same task on multiple scales. The CTAM facilitates the exchange of relevant information across the task features on the same scale, while minimizing an interference between different tasks. The CSAM aggregates useful information from the feature maps with varying resolutions due to different receptive fields in the same task. The two modules are performed in a similar way using the CAM. In this section, we first introduce the CAM and then explain how it is applied in terms of tasks and scales.

**Cross Attention Module (CAM)** The target feature $t_f$ and the source feature $s_f$ are first transformed to a query $Q_t \in R^{H \times W \times d_k}$, a key $K_s \in R^{H \times W \times d_k}$ and a value $V_s \in R^{H \times W \times d_v}$. We use 1×1 Conv-BN-ReLU layer $l_v$ and 1×1 Conv-BN-Softplus layer $l_q$ and $l_k$ for the transformation [7]:

$$Q_t = l_q(t_f) \quad K_s = l_k(s_f) \quad V_s = l_v(s_f). \tag{1}$$

The CAM is calculated by applying the cross-attention and concatenating the target feature as follows:

$$CAM(t_f, s_f) = t_f \oplus CA(t_f, s_f),$$
$$CA_b(t_f, s_f) = \frac{\sum_{a=1}^{L} (Q_{t_b} \cdot K_{s_a}) \cdot V_{s_a}}{\sum_{a=1}^{L} (Q_{t_b} \cdot K_{s_a})}, \tag{2}$$

where $L$ is the number of pixels and $A \oplus B$ means a concatenation operation between $A$ and $B$. $a$ and $b$ denote pixels. Note that $Q_{t_b} \in R^{d_k}$, $K_{s_a} \in R^{d_k}$, and $V_{s_a} \in R^{d_v}$.

**Cross Task Attention Module (CTAM)** The CTAM allows the exchange of relevant information between tasks with minimal interference between them. The CAM is applied by setting task $i$ as the target and the rest as the source in order to transfer useful information of the remaining tasks to the features of task $i$. On the scale $k$, this is repeatedly applied to all tasks from $i = 1, ..., M$ as follows:

$$\tilde{F}_{t_i}^{k} = F_{t_i}^{k} + conv(\mathcal{F}_{\substack{j=1 \\ j\neq i}}^{M} CAM(F_{t_i}^{k}, F_{t_j}^{k})), \qquad (3)$$

where $\mathcal{F}$ means an operation that concatenates all elements from 1 to $M$, except for $j = i$, and $conv$ means $1 \times 1$ convolution operation. After the convolution operation, we also add the original target feature as a residual to preserve the target feature information.

**Cross Scale Attention Module (CSAM)** We use the CSAM to transfer information between different resolution features within the same task. The CAM is applied by setting the $k$-scale feature and the smaller scale feature as the target and source as below:

$$\tilde{F}_{t_i}^{k} = \begin{cases} \mathcal{F}_{l=k+1}^{K} CAM(\tilde{F}_{t_i}^{k}, conv(\tilde{F}_{t_i}^{l})), & k = 1, ..., K-1 \\ \tilde{F}_{t_i}^{k}, & k = K \end{cases} \qquad (4)$$

Here, we set $K$ to 4. $\mathcal{F}$ means the operation that concatenates all elements from $k + 1$ to $K$. The coarsest feature $\tilde{F}_{t_i}^{k}$ is initialized with the CTAM output feature. The feature information smaller than the current scale is propagated into the current scale feature. Finally, we obtain $\tilde{F}_{t_i}^{k}$ by concatenating all the features of the $l$ scale that are useful to the $k$ scale. Unlike the CTAM, the residual layer is not applied as the source and target features are from the same task.

### 2.3. Feature Propagation Module with Attention (FPMA)

In [11], the propagation of the low-resolution feature with a larger receptive field into the higher scale was proven to be effective in enhancing a high-resolution feature. We further boost this feature propagation module by including the self-attention module that can capture the long range dependencies. Also, the output of the CTAM with relevant information from the other tasks are transmitted to the next scale using the FPMA, allowing for the propagation of the task-augmented features across different scales.

**Table 1**: NYUD-v2 results for two tasks (monocular depth estimation and semantic segmentation).

| Model | Depth ↓ | SemSeg ↑ | $\Delta_m$ ↑ |
|---|---|---|---|
| | RMSE | mIoU | |
| Single task baseline | 0.644 | 35.04 | +0.00 |
| Multi task baseline | 0.674 | 35.03 | -2.46 |
| PADNet [6] | 0.624 | 36.72 | +3.94 |
| MTINet [11] | 0.611 | 37.21 | +5.65 |
| ATRC [7] | 0.613 | 40.99 | +10.89 |
| Ours Student | **0.597** | 40.21 | +11.02 |
| Ours Teacher | 0.604 | **41.33** | **+12.07** |

**Table 2**: NYUD-v2 results for three tasks (monocular depth estimation and semantic segmentation, and surface normal estimation).

| Model | Depth ↓ | SemSeg ↑ | Normal ↓ | $\Delta_m$ ↑ |
|---|---|---|---|---|
| | RMSE | mIoU | Mean | |
| Single task baseline | 0.625 | 37.99 | 20.87 | +0.00 |
| Multi task baseline | 0.646 | 35.55 | 21.99 | -5.03 |
| PADNet [6] | 0.640 | 35.75 | 21.28 | -4.41 |
| MTINet [11] | 0.598 | 37.17 | 21.28 | +0.07 |
| ATRC [7] | 0.609 | 37.50 | **20.50** | +1.02 |
| Ours | **0.584** | **40.50** | 20.59 | **+4.82** |

### 2.4. Feature Aggregation

A final feature representation for each task is computed by utilizing the features at multiple scales as:

$$p_{t_i} = conv(\mathcal{F}_{k=1}^{K} U(\tilde{F}_{t_i}^{k})), \qquad (5)$$

where $U$ means an upsampling operation to the scale of a final output $p_{t_i}$. The upsampled features are concatenated, and the $1 \times 1$ convolution operation is performed to infer the final output $p_{t_i}$ for task $i$.

## 3. EXPERIMENTS

### 3.1. Implementation Details and Evaluation Metrics

We used HRNet18 [15] as the backbone network. The performance was measured on the NYUD-v2 [17] and Pascal-Context [18] datasets which are widely used in the multi-task learning. The NYUD-v2 dataset provides 795 training and 654 testing images of indoor scenes. The Pascal-Context dataset consists of 4998 training and 5105 testing images. During training, we resized images to a resolution of $640 \times 480$. We trained the whole network for 100 epochs using Adam optimizer [19] with a learning rate of $10^{-4}$.

For learning the multi-task networks, we performed the semantic segmentation, monocular depth estimation, and surface normal estimation on the NYUD-v2 dataset [17], and performed the semantic segmentation, part segmentation, and surface normal estimation on the Pascal-Context dataset [18], respectively. A cross entropy loss was used for the semantic segmentation and part segmentation, and L1 loss was used for the monocular depth estimation and surface normal. For a performance evaluation, we used the mean intersection over union (mIoU) in the semantic segmentation and part segmentation, the root mean square error (RMSE) in the monocular depth estimation, and the mean angular error (Mean) for the surface normal estimation.

Following [11], we measured the multi-task learning performance $\Delta_m$ which is defined as an average drop in performance per task compared to a single task baseline $b$. If a lower value is better for the performance measure $M_i$ of task $i$, then $l_i = 1$, and otherwise $l_i = 0$. The multi-task learning performance $\Delta_m$ is expressed as follows:

$$\Delta_m = \frac{1}{M}\sum_{i=1}^{M} (-1)^{l_i}(M_{m,i} - M_{b,i})/M_{b,i}. \qquad (6)$$

**Table 5**: Ablation study of our model on the NYUD-v2 dataset. The baseline model is MTINet [11]. CNN with attention indicates using the feature maps augmented by the self-attention module to capture the long-range dependencies. The FPMA denotes the feature propagation module with attention, the CTAM for the cross-task attention module and the CSAM for the cross-scale attention module.

| Model | Methods | | | | Depth Estimation ↓ | | Semantic Segmentation ↑ | $\Delta_m$ ↑ |
|---|---|---|---|---|---|---|---|---|
| | CNN with attention | FPMA | CTAM | CSAM | RMSE | RMSE(log) | mIoU | |
| MTINet [11] | | | | | 0.611 | 0.2086 | 37.21 | +0.00 |
| Ours | ✓ | | | | 0.610 | 0.2086 | 38.80 | +2.21 |
| | ✓ | ✓ | | | 0.609 | 0.2085 | 38.94 | +2.48 |
| | ✓ | ✓ | ✓ | | 0.605 | 0.2077 | 39.14 | +3.08 |
| | ✓ | ✓ | ✓ | ✓ | **0.604** | **0.2064** | **41.33** | **+6.10** |

**Table 3**: Pascal-Context results for two tasks (surface normal estimation and semantic segmentation).

| Model | Normal ↓ | SemSeg ↑ | $\Delta_m$ ↑ |
|---|---|---|---|
| | Mean | mIoU | |
| Single task baseline | 14.87 | 57.33 | +0.00 |
| Multi task baseline | 14.61 | 53.95 | -3.81 |
| PADNet [6] | 14.77 | 54.18 | -2.39 |
| MTINet [11] | 14.55 | 59.08 | +2.59 |
| ATRC [7] | **13.58** | 55.49 | +2.73 |
| Ours | 13.89 | **60.09** | **+5.69** |

**Table 4**: Pascal-Context results for three tasks (surface normal estimation and semantic segmentation, and part segmentation).

| Model | Normal | Semseg | Partseg | $\Delta_m$ ↑ |
|---|---|---|---|---|
| | Mean | mIoU | mIoU | |
| Single task baseline | 14.87 | 57.33 | 60.08 | +0.00 |
| Multi task baseline | 14.95 | 53.79 | 59.45 | -2.56 |
| PADNet [6] | 15.04 | 55.77 | 59.81 | -1.42 |
| MTINet [11] | 14.94 | 59.04 | 61.56 | +1.66 |
| ATRC [7] | **13.71** | 57.94 | 58.08 | +1.84 |
| Ours | 14.71 | **59.10** | **62.47** | **+2.69** |

We used Eq. (6) to compare the single task baseline, multi-task baseline, and the recent methods. The single task baseline model consists of a backbone network (HRNet-18) and a decoder. The multi-task baseline model also consists of a backbone network (HRNet-18) and multiple decoders. For a fair comparison, we re-trained all methods using author-provided codes under the same environment. Our code will be publicly available later.

### 3.2. Evaluation on NYUD-v2 dataset

In Table 1, we compared our model with the recent multi-task learning models using the NYUD-v2 dataset for the two tasks (monocular depth estimation and semantic segmentation). Our model improved the multi-task learning performance by 12.07% over the single task baseline model. In addition, our model outperformed the ATRC [7], which is the state-of-the-art, for the multi-task learning with the semantic segmentation and monocular depth estimation. We also evaluated the performance for the three tasks (monocular depth estimation, semantic segmentation, and surface normal estimation). As shown in Table 2, we achieved the highest performance in the three tasks (+4.82%). The state-of-the-art performance was attained on the monocular depth estimation and semantic segmentation, except for the surface normal estimation which achieves a comparable performance.

### 3.3. Evaluation on Pascal-Context dataset

Table 3 shows a comparison with the recent multi-task learning approaches on the Pascal-Context dataset for the two tasks (semantic segmentation and surface normal estimation). We obtained a 5.69% improvement in the multi-task learning performance when compared to the single task baseline model. In Table 2.4, we additionally compared the multi-task learning performance with the recent models for the three tasks (surface normal estimation, semantic segmentation, and part segmentation), achieving the highest performance improvement of 2.69%.

### 3.4. Ablation Study

In Table 5, we performed an ablation study to prove the effectiveness of our proposed method. The performance evaluation was conducted on the semantic segmentation and monocular depth estimation with the NYUD-v2 dataset. By using the feature maps augmented by the self-attention module ('ATT' in Fig. 1), we achieved a 2.21% improvement over MTINet [11]. The FPMA further improves the performance by 2.48%. The performance gain by the two cross attention modules (CTAM and CSAM) also indicates the effectiveness of the proposed method.

## 4. CONCLUSION

This paper proposed a novel architecture that effectively transfers informative features through different tasks (CTAM) and scales (CSAM) by applying the cross-attention sequentially for scale and task. Our proposed methods have achieved the significant performance improvement over the recent multi-task learning approaches. It is expected that the network design based on the NAS [20, 21] can contribute to improving the performance of the multi-task network. We reserve this as future works.